\documentclass[runningheads]{llncs}
\usepackage{graphicx}
\usepackage{amsmath,amssymb} 
\usepackage{graphicx}
\usepackage{amsmath, amssymb} 
\usepackage{color}
\usepackage{mathtools}
\usepackage{algorithm}
\usepackage{algorithmic}
\usepackage{subcaption}
\usepackage{hyperref}
\usepackage{appendix}

\usepackage{color}

\begin{document}
\pagestyle{headings}
\mainmatter

\def\ACCV20SubNumber{126}  
\newcommand{\algoname}{SDRSAC }
\newcommand{\ttinneriter}{\texttt{inner\_iters}}
\newcommand{\ttit}{\texttt{iter}}
\newcommand{\ttmaxiter}{\texttt{max\_iter}}
\newcommand{\ttbestScore}{\texttt{best\_score}}
\newcommand{\ttbestR}{\texttt{best\_rotation}}
\newcommand{\ttbestT}{\texttt{best\_translation}}

\newcommand{\cM}{\mathcal{M}}
\newcommand{\cI}{\mathcal{I}}
\newcommand{\cS}{\mathcal{S}}
\newcommand{\cD}{\mathcal{D}}
\newcommand{\cP}{\mathcal{P}}
\newcommand{\cQ}{\mathcal{Q}}
\newcommand{\cT}{\mathcal{T}}
\newcommand{\cad}{\mathcal{d}}
\newcommand{\cX}{\mathcal{X}}
\newcommand{\cXh}{\hat{\mathcal{X}}}
\newcommand{\cC}{\mathcal{C}}
\newcommand{\cF}{\mathcal{F}}

\newcommand{\bm}{\mathbf{m}}
\newcommand{\br}{\mathbf{r}}
\newcommand{\bx}{\mathbf{x}}
\newcommand{\bxh}{\hat{\mathbf{x}}}
\newcommand{\bX}{\mathbf{X}}
\newcommand{\bY}{\mathbf{Y}}
\newcommand{\bZero}{\mathbf{0}}
\newcommand{\hbX}{\hat{\mathbf{X}}}
\newcommand{\bS}{\mathbf{S}}
\newcommand{\bs}{\mathbf{s}}
\newcommand{\bp}{\mathbf{p}}
\newcommand{\bq}{\mathbf{q}}
\newcommand{\bD}{\mathbf{D}}
\newcommand{\bd}{\mathbf{d}}
\newcommand{\bA}{\mathbf{A}}
\newcommand{\bR}{\mathbf{R}}
\newcommand{\bt}{\mathbf{t}}
\newcommand{\bH}{\mathbf{H}}
\newcommand{\by}{\mathbf{y}}
\newcommand{\bz}{\mathbf{z}}
\newcommand{\bu}{\mathbf{u}}
\newcommand{\ba}{\mathbf{a}}
\newcommand{\bg}{\mathbf{g}}
\newcommand{\bOnes}{\mathbf{1}}
\newcommand{\bF}{\mathbf{F}}
\newcommand{\bK}{\mathbf{K}}
\newcommand{\bI}{\mathbf{I}}
\newcommand{\tdf}{\tilde{f}}
\newcommand{\tdh}{\tilde{h}}

\newcommand{\bbR}{\mathbb{R}}
\newcommand{\bbE}{\mathbb{E}}
\newcommand{\bbD}{\mathbb{D}}
\newcommand{\bbF}{\mathbb{F}}
\newcommand{\bbFh}{\hat{\mathbb{F}}}
\newcommand{\bmu}{\boldsymbol{\mu}}
\newcommand{\bhr}{\hat{\mathbf{r}}}
\newcommand{\bJ}{\mathbf{J}}
\newcommand{\Nsample}{$N_\text{sample}$}

\newcommand{\kernel}{\psi}

\newcommand{\residual}{\mathbf{r}}
\newcommand{\btheta}{\boldsymbol{\theta}}

\title{Progressive Batching for Efficient \\ Non-linear Least Squares} 
\titlerunning{Progressive Batching for Efficient \\ Non-linear Least Squares }

%
\author{Huu Le\inst{1}\orcidID{0000-0001-7562-7180} \and
Christopher Zach\inst{1}\orcidID{0000-0003-2840-6187} \and Edward Rosten \inst{2} \orcidID{0000-0001-8675-4230}
\and
Oliver J. Woodford\inst{2}\orcidID{0000-0002-4202-4946}
}
\authorrunning{H. Le et al.}
%
\institute{
Chalmers University, Sweden \footnote{This work was partially supported by the Wallenberg AI, Autonomous Systems and Software Program (WASP) funded by the Knut and Alice Wallenberg Foundation.}
\and Snap, Inc., London \& Santa Monica
}

\maketitle

\begin{abstract}
Non-linear least squares solvers are used across a broad range of offline and real-time model fitting problems.
Most improvements of the basic Gauss-Newton algorithm tackle convergence guarantees or leverage the sparsity of the underlying problem structure for computational speedup.
With the success of deep learning methods leveraging large datasets, stochastic optimization methods received recently a lot of attention.
Our work borrows ideas from both stochastic machine learning and statistics, and we present an approach for non-linear least-squares that guarantees convergence while at the same time significantly reduces the required amount of computation.
Empirical results show that our proposed method achieves competitive convergence rates compared to traditional second-order approaches on common computer vision problems, such as image alignment and essential matrix estimation, with very large numbers of residuals.
\end{abstract}

\section{Introduction}
Non-linear least squares (NLLS) solvers~\cite{tingleff2004methods} are the optimizers of choice for many computer vision model estimation and fitting tasks~\cite{hartley2003multiple}, including photometric image alignment~\cite{baker2004lucas}, essential matrix estimation~\cite{hartley2003multiple} and bundle adjustment~\cite{triggs1999bundle}. Fast convergence due to the second-order gradient model, and a simple, efficient implementation due to Gauss' approximation of the Hessian, make it a highly effective tool for these tasks.
Nevertheless, the (non-asymptotic) computational efficiency of these methods can significantly impact the overall performance of a vision system, and the need to run such tasks in real-time, at video frame-rate, for applications such as Augmented Reality, leads to ongoing research to improve NNLS solvers. 

Standard NLLS solvers such as the Gauss-Newton (GN)~\cite{nocedal} or Levenberg-Marquardt (LM) method~\cite{levenberg1944method,marquardt1963algorithm} evaluate all residuals and their Jacobians (first derivatives of the residual function) at every iteration.
Analogous to large-scale machine learning, utilizing all the data available to a problem can therefore substantially and unnecessarily slow down the optimization process.
No improvements in solver efficiency have seen widespread adoption for model fitting, to address this problem. In practice, systems are engineered to pre-select a sparse set of data to avoid it, requiring some tuning on the part of the implementer to ensure that enough data is included for robustness and accuracy, while not too much is included, to keep computation time within budget. These design decisions, made at ``compile time'', do not then adapt to the unknown and variable circumstances encountered at run time.

Inspired by the stochastic methods used to accelerate large-scale optimization problems in machine learning, we introduce a stochastic NLLS optimizer that significantly reduces both computation time and the previously linear relationship between number of residuals and computation time, at no cost to accuracy. Our method has the following novel features:
\begin{itemize}
    \item A stochastic, variable batch size approach for non-linear least squares that can be easily integrated into existing NLLS solvers for model fitting applications.
    \item A statistical test to determine the acceptance of an update step computed from a batch, without evaluating all residuals, that also progressively increases the batch size.
    \item Guaranteed convergence to a local minimum of the original problem, since all residuals are automatically included in the final iterations.
\end{itemize}
By adjusting the batch size at run time, according to a reasoned, statistical test, our algorithm is able to invest more computational resources when required, and less when not. This avoids the need to tightly tune the number of residuals at compile time, and can therefore lead to more accurate performance as a result.
 
We evaluate our method on a number of two-view geometry problems involving geometric and photometric residuals, such as essential matrix estimation and homography fitting, with promising results\footnote{Our source code is available at \url{https://github.com/intellhave/ProBLM}}.
In particular, we empirically show that our new approach has much faster convergence rates compared to conventional approaches.

\section{Related Work}
\subsection{Non-linear least squares}
Fully second-order optimizers, such as NLLS methods, benefit from both the automatic choice of the step size and from modelling the dependency between variables, which results in faster convergence (rates) than first order and even quasi-Newton methods.
Gauss replaced the Hessian of Newton's method with an approximation for least squares costs in 1809, creating the original NLLS solver, the Gauss-Newton method~\cite{nocedal}.
Since the Gauss-Newton method does not guarantee convergence, the Levenberg-Marquardt algorithm~\cite{levenberg1944method,marquardt1963algorithm} extends Gauss-Newton with back-tracking (i.e.\ conditional acceptance of new iterates) and diagonal damping of the (approximate) Hessian.
More recently, a further modification of the Gauss-Newton method, variable projection (VarPro~\cite{golub1973varpro,okatani2007wiberg,hong2017varpro}), has been proposed for a particular class of separable problems, resulting in wider convergence basins to reach a global solution.

Since many model fitting tasks are solved using NLLS, several acceleration techniques have been developed to address problems of scale and real-time operation. However, these techniques are not generic, but instead exploit task specific properties. For example, certain image alignment tasks have been accelerated by the inverse compositional algorithm~\cite{baker2004lucas}, which computes Jacobians once only, in the reference frame, or by learning a hyperplane approximation to the pseudo-inverse~\cite{jurie2002hyperplane}.
Truncated and inexact Newton methods~\cite{nocedal} typically use iterative solvers such as conjugate gradients (CG) to approximately solve the linear system. These can converge in less time on larger scale problems, such as bundle adjustment (BA), especially when efficient, BA specific preconditioners are used~\cite{agarwal2010bundle}.
In a framework such as RANSAC~\cite{fischler1981random}, convergence for each subproblem is not required so significant speedups are available by allowing a reduced probability of convergence~\cite{rosten_2010_improved}.

Huge gains in efficiency are available if the problem exhibits a conditional independence  structure.
Bundle adjustment and related bipartite problem instances use techniques such as the Schur complement~\cite{triggs1999bundle} (or more generally column reordering~\cite{davis2004algorithm}) to reduce the size of the linear system to be solved in each iteration.
Other linear systems have linear time solvers: Kalman smoothing~\cite{kalman1960filter} is a special case of using belief propagation~\cite{pearl1982reverend} to solve linear least squares and 
such techniques can be applied to non-linear, robust least squares over tree structures~\cite{drummond2001realtime}.


\subsection{Stochastic methods}
Stochastic first order methods~\cite{robbins1951stochastic,kiefer1952stochastic}, which are now common for large-scale optimization in machine learning, compute approximate (first order) gradients using subsets of the total cost, called batches, thereby significantly accelerating optimization. The randomness of the method (and the intermediate solutions) requires certain provisions in order to obtain guarantees on convergence. Due to their stochastic nature these methods empirically return solutions located in ``wide'' valleys~\cite{wilson2003general,keskar2016large}.

In addition to first-order methods stochastic second-order one have also been investigated (e.g.~\cite{byrd2016stochastic,bollapragada2018progressive,zhao2018stochastic,curtis2019fully}). One main motivation to research stochastic second-order methods is to overcome some shortcomings of stochastic first-order methods such as step size selection by introducing curvature (2nd-order) information of the objective.
Due to the scale of problems tackled, many of these proposed algorithms are based on the L-BFGS method~\cite{nocedal1980updating,liu1989limited}, which combines line search with low-rank approximations for the Hessian.
The main technical difference between stochastic first-order and second-order methods is that the update direction in stochastic gradient methods is an unbiased estimate of the true descent direction, which is generally not the case for stochastic second-order updates.
Convergence of stochastic methods relies on controlling the variance of the iterates, and consequently requires e.g.\ diminishing step sizes in stochastic first order methods~\cite{robbins1951stochastic,bottou2018optimization} or e.g.\ shrinking trust region radii~\cite{curtis2019fully} for a second order method. Without diminishing variances of the iterates, it is only possible to return approximate solutions~\cite{tran2020stochastic}.
Hence, using a proper stochastic method to minimize an NNLS instance over a finite number of residuals will never fully benefit from the fast convergence rates of 2nd order methods.

A number of recent proposals for stochastic methods use batches with an adaptive (or dynamically adjusted) batch size.
This is in particular of interest in the second-order setting, since increasing the batch is one option to control the variance of the iterates (another option being reducing the step size).
\cite{byrd2012sample} and~\cite{bollapragada2018adaptive} propose schemes to adjust the batch size dynamically, which mostly address the first-order setup.
Bollapragada et al.~\cite{bollapragada2018progressive} propose a randomized qausi-Newton method, that is designed to have similar benefits as the stochastic gradient method. The method uses progressively larger sets of residuals, and the determination of the sample size is strongly linked with the underlying L-BFGS algorithm.
A trust region method with adaptive batch size is described in~\cite{mohr2019adaptive}, where the decision to accept a new iterate is based on the full objective (over all residuals), rather than the batch subset, limiting the benefits of this method. Similarly,~\cite{agarwal2017second} and~\cite{pilanci2017newton} propose stochastic second-order methods that build on stochastic approximations of the Hessian but require computation of the full gradient (which was later relaxed in~\cite{roosta2019sub}).



The stochastic approaches above target optimizations with a large number of model parameters, which each depend on a large number of residuals.
The scale of such problems does usually not permit the direct application of standard NLLS solvers.
In many model fitting problems in computer vision, the number of variables appearing in the (Schur-complement reduced) linear system is small, making NLLS a feasible option.
L-BFGS~\cite{byrd1995limited} (and any quasi-Newton method) has a disadvantage compared to Gauss-Newton derivatives for NNLS problem instances, since the NNLS objective is near quadratic when close to a local minimum. Thus, L-BFGS is not favoured in these applications, due to its slower convergence rate than NLLS (which is also empirically verified in Section~\ref{sec:experiments}).
We note that the progressive batching of residuals introduced in this work can in principle also be applied to any local optimization method that utilizes back-tracking to conditionally accept new iterates.

\section{Background}
\subsection{Problem Formulation}
Our work tackles the following NLLS problem:
\begin{align}
    \min_{\btheta} \sum\nolimits_{i=1}^N f_i (\btheta),~~~\textrm{where}~~~ f_i(\btheta) = \|\br_i (\btheta)\|_2^2
    \label{eq:nonlin_lsq}
\end{align}
where $\btheta \in \bbR^d$ is the vector containing the desired parameters, $N$ specifies the number of available measurements (residuals), and $\br_i :\bbR^d \mapsto \bbR^p$ is the function that computes the residual vector (of length $p$) of the $i$-th measurement.

It is worth noting that, while we start by introducing the standard NLLS formulation as per~\eqref{eq:nonlin_lsq}, our proposed method is directly applicable to robust parameter estimation problems through the use of methods such as Iteratively Reweighted Least Squares (IRLS)~\cite{holland1977irls}, since each step of IRLS can be re-formulated as a special instance of~\eqref{eq:nonlin_lsq}. More specifically, with the use of a robust kernel $\psi$ to penalize outlying residuals, the robust parameter estimation is defined as, 
\begin{align}
    \min_{\btheta} \sum\nolimits_{i=1}^N g_i(\btheta),~~~\textrm{where}~~~ g_i(\btheta)= \psi(\|\br_i (\btheta)\|).
    \label{eq:robust_nonlin_lsq}
\end{align}

\subsection{Levenberg-Marquardt Method}
\label{sec:lm}
While our progressive batching approach is applicable to any second-order algorithm, here we employ Levenberg-Marquardt (LM) as the reference solver, as it is the most widely used method in a number of computer vision problems. In this section we briefly review LM and introduce some notations which are also later used throughout the paper.

At the $t$-th iteration, denote the current solution as $\btheta^{(t)}$ and $\{\br_i(\btheta^{(t)})\}_{i=1}^N$ as the set of residual vectors for all measurements. The LM algorithm involves computing the set of Jacobians matrices $\{\bJ^{(t)}_i \in \bbR^{p\times d}\}$, where, using $\br^{(t)}_i$ as shorthand for $\br_i(\btheta^{(t)})$,
\begin{equation}
    \bJ^{(t)}_i = \begin{bmatrix}
    \frac{\partial \br^{(t)}_i}{\partial \btheta^{(t)}_1} 
    & \frac{\partial \br^{(t)}_i}{\partial \btheta^{(t)}_2}
    & \dots \frac{\partial \br^{(t)}_i}{\partial \btheta^{(t)}_d}
    \end{bmatrix}.
\end{equation}
Based on the computed Jacobian matrices and the residual vectors, the gradient vector $\bg^{(t)}$ and approximate Hessian matrix $\bH^{(t)}$ are defined as follows:
\begin{equation}
    \bg^{(t)} \coloneqq \sum\nolimits_{i=1}^N (\bJ^{(t)}_i)^T\br_i^{(t)}, ~~~~~~~ \bH^{(t)} \coloneqq \sum\nolimits_{i=1}^N (\bJ^{(t)}_i)^T \bJ^{(t)}_i,
    \label{eq:lm_grad_hess_def}
\end{equation}
where $(.)^T$ denotes the matrix transpose operation. Given $\bg^{(t)}$ and $\bH^{(t)}$, LM computes the step $\Delta \btheta$ by solving
\begin{equation}
    \Delta\btheta^{(t)} \leftarrow -(\bH^{(t)} + \lambda \bI)^{-1} \bg^{(t)},
    \label{eq:lm_step}
\end{equation}
where $\lambda$ is the damping parameter that is modified after each iteration depending on the outcome of the iteration. In particular,
if the computed step leads to a reduction in the objective function, i.e., $f(\btheta^{(t)} + \Delta\btheta^{(t)}) < f(\btheta^{(t)})$, the step is accepted and $\btheta$ is updated by setting $\btheta^{t+1} \leftarrow \btheta^{t} + \Delta\btheta$, while the damping $\lambda$ is decreased by some factor. On the other hand, if $f(\btheta^{(t)} + \Delta\btheta^{(t)}) \ge f(\btheta^{(t)})$, $\Delta\btheta^{(t)}$ is rejected, $\lambda$ is increased accordingly and \eqref{eq:lm_step} is recomputed; this is repeated until the cost is reduced. 

\section{Proposed Algorithm}

In this section, we describe our proposed algorithm that is computationally cheaper than conventional second-order approaches. The main idea behind our algorithm is that, instead of computing the Jacobians of all residuals (as described in Sec.~\ref{sec:lm}), we utilize only a small fraction of measurements to approximate the gradient and Hessian, from which the update step is computed.

Let $\cS^{(t)} \subseteq \{1 \dots N\}$ denote the subset of residual indices that are used at the $t$-th iteration (note that the set of subsampled indices can remain the same throughout many iterations, i.e., $\cS^{t_0} = \dots = \cS^{t} (t_0 \le t) $). For the ease of notation, we use $\cS$ to specify the subset being used at the current iteration. Given a subset $\cS$, analogous to~\eqref{eq:lm_grad_hess_def}, we use the notation ${\bg}_{\cS}$ and ${\bH}_{\cS}$ to denote the approximate gradient and Hessian obtained by only using residuals in the subset $\cS^{(t)}$, i.e.
\begin{align}
    {\bg}^{(t)}_{\cS} \coloneqq \sum\nolimits_{i \in \cS^{(t)}} \bJ^{(t)}_i \br_i^{(t)} & & {\bH}^{(t)}_{\cS} \coloneqq \sum\nolimits_{i \in \cS^{(t)}} (\bJ^{(t)}_i)^T \bJ^{(t)}_i.
    \label{eq:subset_grad_hessian}
\end{align}
We also define the subset cost $f^{(t)}_{\cS}$ as $f^{(t)}_{\cS} \coloneqq \sum_{i \in \cS} f_i(\btheta)$.  Similar to~\eqref{eq:lm_step}, the approximate update step is denoted by $\Delta \hat{\btheta}_{\cS^{(t)}}$, and is computed by 
\begin{equation}
    \Delta {\btheta}^{(t)}_{\cS} \leftarrow -({\bH}^{(t)}_{\cS} + \lambda \bI)^{-1}{\bg}^{(t)}_{\cS}.
    \label{eq:slm_step}
\end{equation}
Note that depending on the characteristic of $\cS^{(t)}$, ${\bg}^{(t)}_{\cS}$ and ${\bH}_{\cS}^{(t)}$ can be very far from the true $\bg^{(t)}$ and $\bH^{(t)}$, respectively. As a consequence, the update step $\Delta \btheta^{(t)}_{\cS}$ computed from~\eqref{eq:slm_step}, despite resulting in a reduction in $f^{(t)}_{\cS}$, may lead to an increase in the original cost $f^{(t)}$. However, as the number of measurements can be very large, computing the whole set of residuals at each iteration to determine the acceptance of $\Delta {\btheta}_{\cS^{(t)}}$ is still very costly. Hence, we employ statistical approaches. Specifically, we only accept $\Delta \btheta^{(t)}_{\cS}$ if, with a probability not less than $1-\delta$ ($0 < \delta < 1$), a reduction in $f^{(t)}_{\cS}$ also leads to a reduction in the true cost $f^{(t)}$. More details are discussed in the following sections.

\subsection{A Probabilistic Test of Sufficient Reduction}
\label{sec:prob_test}
We now introduce a method to quickly determine if an update step $\Delta {\btheta}^{(t)}_{\cS}$ obtained from~\eqref{eq:slm_step} also leads to a sufficient reduction in the original cost with a high probability. To begin, let us define $\btheta^{(t+1)} \coloneqq \btheta^{(t)} + \Delta {\btheta}^{(t)}_{\cS}$, and denote by  
\begin{align}
    X_i = f_i(\btheta^{(t+1)}) - f_i(\btheta^{(t)})
\end{align}
the change of the $i$-th residual. We convert $X_i$ to a random variable by drawing the index $i$ from a uniform distribution over $\{1,\dotsc,N\}$. Taking $K$ random indices (which form the subset $\cS$) yields the random variables $(Y_1,\dotsc,Y_K)$, where
\begin{align}
    Y_k = X_{i_k} \qquad i_k \sim U\{1,\dotsc,N\}.
\end{align}
We can observe that the expectation 
\begin{align}
    \mathbb{E}[Y_k] = f(\btheta^{(t+1)}) - f(\btheta^{(t)})
    = \sum\nolimits_i \left( f_i(\btheta^{(t+1)}) - f_i(\btheta^{(t)}) \right),
\end{align}
represents the total change of the true objective, and at each iteration we are interested in finding $\btheta^{(t+1)}$ such that $\mathbb{E}[Y_k] < 0$. To obtain a lower bound for the random variables (which will be useful for the test introduced later), we can clamp $Y_k$ to a one-sided range $[a,\infty)$ by introducing $Z_k := \max(a, Y_k)$. It can be noted that $\mathbb{E}[Y_k] \le \mathbb{E}[Z_k]$ and therefore $P(\mathbb{E}[Y_k] \ge 0) \le P(\mathbb{E}[Z_k] \ge 0)$, hence we can safely use $\mathbb{E}[Z_k]$ as a proxy to evaluate $\mathbb{E}[Y_k]$. 

We introduce $S_K := \sum_{k=1}^K Z_k$, representing (an upper bound to) the observed reduction, i.e. $S_K \ge f_{\cS}(\btheta^{(t+1)}) - f_{\cS}(\btheta^{(t)}) $. Recall that, during optimization, $S_K$ is the only information available to the algorithm, while our real information of interest is the expectation $\mathbb{E}[Z_k]$. Therefore, it is necessary to establish the relation between $\mathbb{E}[Z_k]$ and $S_K$. Assume that $S_K < 0$ (i.e., the update step leads to a reduction in the observed cost), given a probability $0 < \delta < 1$, and a scalar $0 \le \alpha < 1$, we are interested in the following criterion,
\begin{align}
    P\left( \mathbb{E}[S_K] \le \alpha S_K \right) \ge 1 - \delta,
    \label{eq:prob_reduction}
\end{align}
indicating whether the true cost is also reduced by at least a fraction $\alpha$ of the observed reduction $S_K$ with probability $1-\delta$. Using Hoeffding's inequality~\cite{hoeffding1994probability}%
\footnote{Hoeffding's inequality is one of the main tools in statistical learning theory, but has seen limited use in computer vision so far (e.g.~\cite{cohen2015likelihood}).}%
, we obtain
\begin{align}
        P(\mathbb{E}[S_K] \ge \alpha S_K) &= P\left( S_K - \mathbb{E}[S_K] \le (1-\alpha)S_K \right)
        \le \exp\left( -\tfrac{2(1-\alpha)^2 S_K^2}{K(b-a)^2} \right),
        \label{eq:prob_reduction_inv}
\end{align}
where $b \in \bbR$ is the upper bound of the random variables $Z_k$ ($Z_k \le b \; \forall k)$. While the lower bound $a$ can be freely chosen, computing $b$ is often more involved in practice (we will discuss several options to choose $b$ in the following section).%
\footnote{Note that these bounds may depend on the current iteration, hence $a$ and $b$ should be understood as $a^{(t)}$ and $b^{(t)}$.}

In order for~\eqref{eq:prob_reduction} to hold, we require the r.h.s. of~\eqref{eq:prob_reduction_inv} to be upper-bounded by a user-specified confidence level $\delta \in (0,1)$, i.e.,
        $\exp\left( -\frac{2(1-\alpha)^2 S_K^2}{K(b-a)^2} \right) \le \delta,$
which leads to the condition
\begin{align}
    S_K \le - \frac{b-a}{1-\alpha} \cdot \sqrt{\frac{-K\log\delta}{2}} \qquad \big( \le 0 \big).
    \label{eq:S_K_condition_sufficient}
\end{align}
Thus, if the condition~\eqref{eq:S_K_condition_sufficient} is satisfied, we can confidently accept the step computed from the subset $\cS$.  More specifically, based on $S_K$, the following steps are applied for the LM iterations on the subset:
\begin{enumerate}
    \item $S_K \ge 0$: increase $\lambda$ (e.g.\ $\lambda \leftarrow 10\lambda$), since the LM step was not successful for even the optimized function (the subsample version of the true objective).
    \item $S_K \le 0$ but Eq.~\eqref{eq:S_K_condition_sufficient} is not satisfied: increase the sample set to $K^+$, $\lambda$ remains unchanged.
    \item $S_K \le 0$ but Eq.~\eqref{eq:S_K_condition_sufficient} is satisfied: decrease $\lambda$ (e.g.\ $\lambda \leftarrow \lambda/10$)
\end{enumerate}

Note that Hoeffding's inequality also holds for sample sets without replacement. This means that indices $i_k$ can be unique and obtained by random shuffling the residuals at the beginning of the algorithm. Let $\pi$ be a random permutation of $\{1,\dotsc,N\}$. Then $\cS_K$ is given by $\cS_K = \{ \pi(k) : k = 1,\dotsc, K \}$.
Thus, it is not necessary to draw batches at every iteration, which drastically reduces the variance of the iterates $\btheta^{(t)}$.
Further, using slightly more general versions of Hoeffding's inequality allows residual specific upper and lower bounds $[a_i,b_i]$, which can be useful especially when residuals can be grouped (e.g.\ into groups for the data terms and for a regularizer).

\subsection{Bounding the change of residuals}
\label{sec:bounding}
\subsubsection{Lower bound $a$} Observe that, both the l.h.s. and r.h.s. of the criterion~\eqref{eq:S_K_condition_sufficient} depend on the lower bound $a$. Due to the fact that $\mathbb{E}[Y_k] \le \mathbb{E}[Z_k]$, the condition~\eqref{eq:S_K_condition_sufficient} is valid for any choices of $a < 0$. One fast option to search for $a$ is to successively select the observed reductions $Y_k$ as values for $a$, and test whether the condition~\eqref{eq:S_K_condition_sufficient} is satisfied for any of them.

\subsubsection{Upper bound $b$}
While choosing the upper bound $b$ used in~\eqref{eq:S_K_condition_sufficient} is generally hard, in practice it can be approximated using several options:
\begin{enumerate}
    \item Each $f_i$ has range $[0,\bar f]$. This is the case e.g.\ when all $\br_i$ are continuous and the domain for $\btheta$ is compact. It is also the case when $f_i$ are robustified, i.e.\ $f_i(\btheta) = \psi(\|\br_i(\btheta)\|)$, where $\psi:\mathbb{R}_{\ge 0} \to [0,1]$ is a robust kernel (such as the Geman-McClure or the Welsch kernels). If the upper bound $\bar f$ for each $f_i$ is known, then $b$ in Eq.~\eqref{eq:S_K_condition_sufficient} is given by $b=\bar y$ (since the worst case increase of a term $f_i$ is from~0 to $\bar y$).
    \item Each $f_i$ is Lipschitz continuous with constant $L_f$. In this case we have
        $| f_i(\btheta) - f_i(\btheta')| \le L_f \| \btheta - \btheta'\|$,
    in particular for $\btheta=\btheta^{(t)}$ and $\btheta=\btheta^{(t+1)}$. This implies that $|f_i(\btheta^{(t+1)}) - f_i(\btheta^{(t)})| \le L_f \| \btheta^{(t+1)} - \btheta^{(t)} \|$ for all $i$, and $b$ in Eq.~\eqref{eq:S_K_condition_sufficient} is therefore given by $b=L_f \|\btheta^{(t+1)}-\btheta^{(t)}\|$. This computation of $b$ can be extended straightforwardly if all $f_i$ are H\"older continuous.
\end{enumerate}
In our experiments, we test our algorithm on both robustified and non-robustified problems. In order to approximate $L_f$ for non-robustified cases, we propose to use the maximum change in sampled residuals, i.e.,
\begin{align}
    L_f = \max_{i \in \cS} \frac{|f_i(\btheta^{(t+1)}) - f_i(\btheta^{(t)})|}{\| \btheta^{(t+1)} - \btheta^{(t)} \|}.
    \label{eq:approximate_Lf}
\end{align}

\subsection{Determining New Sample Sizes}
At any iteration, when the condition~\eqref{eq:S_K_condition_sufficient} fails, the sample size $K$ is increased to $K^+ > K$, and the algorithm continues with an extended subset $\cS^+$ with $|\cS^+| = K^+$. In this work, we approximate $K^+$ as follows:
we use the estimate $\hat S_{K^+} = K^+ S_K/K$ for $S_{K^+}$ and choose $K^+$ such that the condition Eq.~\eqref{eq:S_K_condition_sufficient} is satisfied for our estimate $\hat S_{K^+}$. After simplification we obtain
\begin{align}
  K^+ = -\frac{K^2 (b-a)^2 \log\delta}{2S_K^2 (1-\alpha)^2}.
\end{align}
If we introduce $\tilde{\delta} := \exp(2S_K^2/(K(b-a)^2))$ as the confidence level such that $P(\mathbb{E}[S_K] \ge 0) \le \tilde{\delta}$ under the observed value $S_K$, then $K^+$ can be stated as $K^+ = K\, \frac{\log \delta}{\log \tilde{\delta}}$.
If $K^+ > N$, then the new batch size is at most $N$. In summary, $K^+$ is given by
\begin{align}
  \label{eq:K_increase}
  K^+ = \min\left\{ N, \left\lceil -\frac{K^2 (b-a)^2 \log\delta}{2S_K^2 (1-\alpha)^2} \right\rceil \right\}.
\end{align}
\subsection{Relaxing the Condition~\eqref{eq:S_K_condition_sufficient}}
If $T_S$ iterations of LM steps on subsampled residuals are applied, then the probability that \emph{all} of these iterations led to a decrease of the full objective is given by $(1-\delta)^{T_S}$. Asking for all steps to be true descent steps can be very restrictive, and makes the condition \eqref{eq:S_K_condition_sufficient} unnecessarily strict.\!\!\footnote{If we allow a ``failure'' probability $\eta_0$ for only increasing steps, then $\delta$ is given by $\delta = 1-\sqrt[T_S]{(1-\eta_0)}$. E.g., setting $T_S=100$ and $\eta_0=10^{-4}$ yields $\delta\approx 10^{-6}$.}
In practice one is interested that (i) most iterations (but not necessarily all) lead to a descent of the true objective, and that (ii) the true objective is reduced at the end of the algorithm (or for a number of iterations) with very high probability.
Let $t_0$ and $t$ be two iterations counters $t_0 < t$ such that the sample sets are constant, $\cS^{(t_0)} = \ldots = \cS^{(t)}$.
Let $T_S=t-t_0+1$ be the number of successful LM iterations, that use the current sample set $\cS^{(t)}$, and introduce the total observed reduction of the sampled cost after $T$ (successful) iterations,
\begin{align}
    U_K^{(t_0,t)} := \sum\nolimits_{r=t_0}^t S_K^{(r)},
\end{align}
and recall that $S_K = U_K^{(t_0,t_0)}$. Let the current iteration be a successful step (leading to a reduction of the sampled objective $f_{\cS_K^{(t)}}$).
With the introduction of $U_k$, following the same reasoning as introduced in Sec.~\ref{sec:prob_test}, our relaxed criterion reads:
\begin{align}
    \label{eq:relaxed_S_K_sufficient}
    U_K^{(t_0,t)} &\le - \frac{1-\alpha}{b-a} \sqrt{\frac{-K \log\delta}{2}}
\end{align}
If the above criterion (with $\alpha \in (0,1)$) is not satisfied, then with probability $\eta \in [0,1)$ the step is temporarily accepted (and $\lambda$ reduced). With probability $1-\eta$ the step is rejected and the sample size is increased. The rationale is that allowing further iterations with the current sample set may significantly reduce the objective value. If the condition~\eqref{eq:relaxed_S_K_sufficient} is never satisfied for the current sample set $\cS^{(t_0)}$, then the expected number of ``wasted'' iterations is $1/(1-\eta)$ (using the properties of the geometric series).

\subsection{The complete algorithm}

We illustrate the complete method in Alg.~\ref{alg:PROBLM}, which essentially follows a standard implementation of the Levenberg-Marquardt method.
One noteworthy difference is that the implementation distinguishes between three scenarios depending on the reduction gain $S_K$ (failure step, success step and insufficient step).
For clarity we describe the basic (non-relaxed) variant of the method, and refer to the supplementary material for the implementation based on the relaxed test (Eq.~\eqref{eq:relaxed_S_K_sufficient} and the details of estimating the lower bound $a$.
In the experiments we refer to our algorithm using the acronym \emph{ProBLM} (\textbf{Pro}gressive \textbf{B}atching \textbf{LM}).

\begin{algorithm}[t]
  \centering
  \caption{Stochastic Levenberg-Marquardt}
\label{alg:PROBLM}
\begin{algorithmic}[1]                   
  \REQUIRE Initial solution $\btheta^{(0)}$, initial batch size $K_0$, maximum iterations \texttt{max\_iter}
  \REQUIRE Confidence level $\delta \in (0,1)$, margin parameter $\alpha \in [0,1)$
  \STATE Randomly shuffle the residuals $\{f_i\}$ and initialize $t \gets 0$, $K \gets K_0$
  \WHILE{$t<$ \texttt{max\_iter} and a convergence criterion is not met}
  \STATE $\cS^{(t)} \gets \{1, \dotsc, K\}$
  \STATE Compute $\bg_{\cS^{(t)}}$ and $\bH_{\cS^{(t)}}$
  \begin{align}
    {\bg}_{\cS^{(t)}} \coloneqq \sum\nolimits_{i \in \cS^{(t)}} \bJ^{(t)}_i \br_i^{(t)} & &  {\bH}_{\cS^{(t)}} \coloneqq \sum\nolimits_{i \in \cS^{(t)}} (\bJ^{(t)}_i)^T \bJ^{(t)}_i.
  \end{align}
  and solve
  \begin{align}
    \Delta {\btheta}^{(t)} \gets \left( {\bH}_{\cS^{(t)}} + \lambda \bI \right)^{-1}{\bg}_{\cS^{(t)}} & & \btheta^{(t+1)} \gets \btheta^{(t)} + \Delta {\btheta}^{(t)}
  \end{align}
  \STATE Determine current lower and upper bounds $a$ and $b$, and set
  \begin{align}
    S_K \gets \sum\nolimits_{i\in \cS^{(t)}} \max\left\{ a, \left( f_i(\btheta^{(t+1)}) - f_i(\btheta^{(t)}) \right) \right\}.
  \end{align}
  \IF{$S_K \ge 0$}
  \STATE $\btheta^{(t+1)} \gets \btheta^{(t)}$ and $\lambda \gets 10\, \lambda$ \hfill $\vartriangleright$ Failure step
  \ELSIF{$S_K$ satisfies Eq.~\eqref{eq:S_K_condition_sufficient}}
  \STATE $\lambda \gets \lambda/10$ \hfill $\vartriangleright$ Success step
  \ELSE
  \STATE $\btheta^{(t+1)} \gets \btheta^{(t)}$ and increase $K$ using Eq.~\eqref{eq:K_increase}
  \ENDIF
  \STATE $t \gets t+1$
  \ENDWHILE
  \RETURN $\btheta^{(t)}$
\end{algorithmic}
\end{algorithm}

\subsection{Convergence}
When \texttt{max\_iter}$\to\infty$, then the convergence properties of the algorithm are the same as for the regular Levenberg-Marquardt method:
if a sample set $\cS^{(t)}$ remains constant for a number of iterations, the method eventually approaches a stationary point of $f_{\cS^{(t)}}$ leading to diminishing reductions $S_K$.
Consequently, Eq.~\eqref{eq:S_K_condition_sufficient} will not hold after a finite number of iterations, and the batch size strictly increases until $K^+=N$ is reached.

\section{Experiments}
\label{sec:experiments}

We choose dense image alignment (with homography model and photometric errors), and essential matrix estimation (with geometric errors) to evaluate the performance our proposed algorithm. Experiments for bundle adjustment can be found in the supplementary. The image pairs used throughout our experiments are obtained from a variety of publicly available datasets, including the ETH3D,\!\!\footnote{\url{https://www.eth3d.net/datasets}} EPFL Multi-view stereo\footnote{\url{https://www.epfl.ch/labs/cvlab/data/data-strechamvs/}} and AdelaideMRF\footnote{\url{https://tinyurl.com/y9u7zmqg}} dataset~\cite{wong2011dynamic}. In this section, we focus on presenting representative results that highlight the performance of our approach. More detailed results and studies of parameters are provided in the supplementary material.
Two types of problems are tested in our experiments:
\begin{itemize}
    \item Standard NLLS (Problem~\eqref{eq:nonlin_lsq}): To test this type of problem, we perform dense homography estimation with photometric errors, and essential matrix refinement using sparse key points (outliers are assumed to be rejected by a pre-processing step, e.g.\ using RANSAC~\cite{fischler1981random}). 
    \item Problems with robustified residuals: We also investigate the performance of our approach on model fitting problems with robust kernels (Problem~\eqref{eq:robust_nonlin_lsq}). The essential matrix estimation on a sparse set of putative correspondences (containing outliers) is performed, where the outliers are directly discarded during the optimization process by applying a robust kernel $\psi$ to the residuals (in contrast to the previous experiments where RANSAC is used to discard outliers). We choose $\psi$ to be the smooth truncated least squares kernel,
    \begin{equation}
        \psi(r) = 
        \tfrac{\tau^2}{4} \left( 1 - \max\{ 0, 1-r^2/\tau^2 \}^2 \right)
        \label{eq:truncated_lsq}
    \end{equation}
    where $\tau$ is the inlier threshold. In this case, the upper bound $b$ on the residual changes that is used in Eq.~\eqref{eq:S_K_condition_sufficient} is $\frac{1}{4}\tau^2$.
\end{itemize}
The standard LM algorithm is used as the baseline to assess the performance of our proposed approach. In addition, we also compared our method against L-BFGS. All algorithms are implemented in C++ and executed on an Ubuntu workstation with an AMD Ryzen 2950X CPU and 64GB RAM. We employ the open-source OpenCV library\footnote{\url{https://github.com/opencv/opencv}} for pre-processing tasks such as SIFT feature extraction and robust fitting with RANSAC. We set  $\delta$ to $0.1$ and $\alpha$ to $0.9$ in all experiments. The initial sample size ($K_0$) is set to  $0.1N$ ($N$ is the number of total measurements). All the experiments use the relaxed version as shown in Eq.~\eqref{eq:relaxed_S_K_sufficient}, where the parameter $\eta$ is set to $0.5$. A comparison between Eq.~\eqref{eq:S_K_condition_sufficient} and its relaxed version is provided in the supplementary material.

\subsection{Dense Image Alignment with Photometric Errors}
This problem is chosen to demonstrate the efficiency of our proposed method as it often requires optimizing over a very large number of residuals (the number of pixels in the source image). In particular, given two images $I_1$ and $I_2$, the task is to estimate the parameters $\btheta \in \bbR^d$ that minimize the photometric error,
\begin{equation}
    \min_{\btheta} \sum\nolimits_{\bx \in I_1} \|I_1(\bx) - I_2(\pi(\bx, \btheta))\|^2,
\end{equation}
where $\bx$ represents the pixel coordinates, and $\pi(\bx, \btheta)$ is the transform operation of a pixel $\bx$ w.r.t. the parameters $\btheta$. In this experiment $\pi$ is chosen to be the homography transformation, thus $\btheta \in \bbR^8$ (as the last element of the homography matrix can be set to $1$).
When linearizing the residual we utilize the combined forward and backward compositional approach~\cite{malis2004improving} (which averages the gradient contribution from $I_1$ and $I_2$), since this is more stable in practice and therefore a more realistic scenario.

We select six image pairs from the datasets (introduced above) to test our method (results for more image pairs can be found in the supplementary materials). Fig.~\ref{fig:dense_homography_results} shows the optimization progress for the chosen image pairs, where we plot the evolution of the objectives vs. the run times for our method and conventional LM. We also compare the results against L-BFGS, where it can clearly be observed that L-BFGS performs poorly for this particular problem (note that for image pairs where L-BFGS does not provide reasonable solutions, we remove their results from the plots).

As can be observed from Fig.~\ref{fig:dense_homography_results}, ProBLM achieves much faster convergence rates compared to LM. Moreover, Fig.~\ref{fig:dense_homography_results} also empirically shows that our proposed method always converges to the same solutions as LM, thanks to our efficient progressive batching mechanism.
Due to the non-linearity of the underlying problem this is somewhat surprising, since the methods will follow different trajectories in parameter space.

\begin{figure}
    \centering
    \includegraphics[width = 0.32\textwidth]{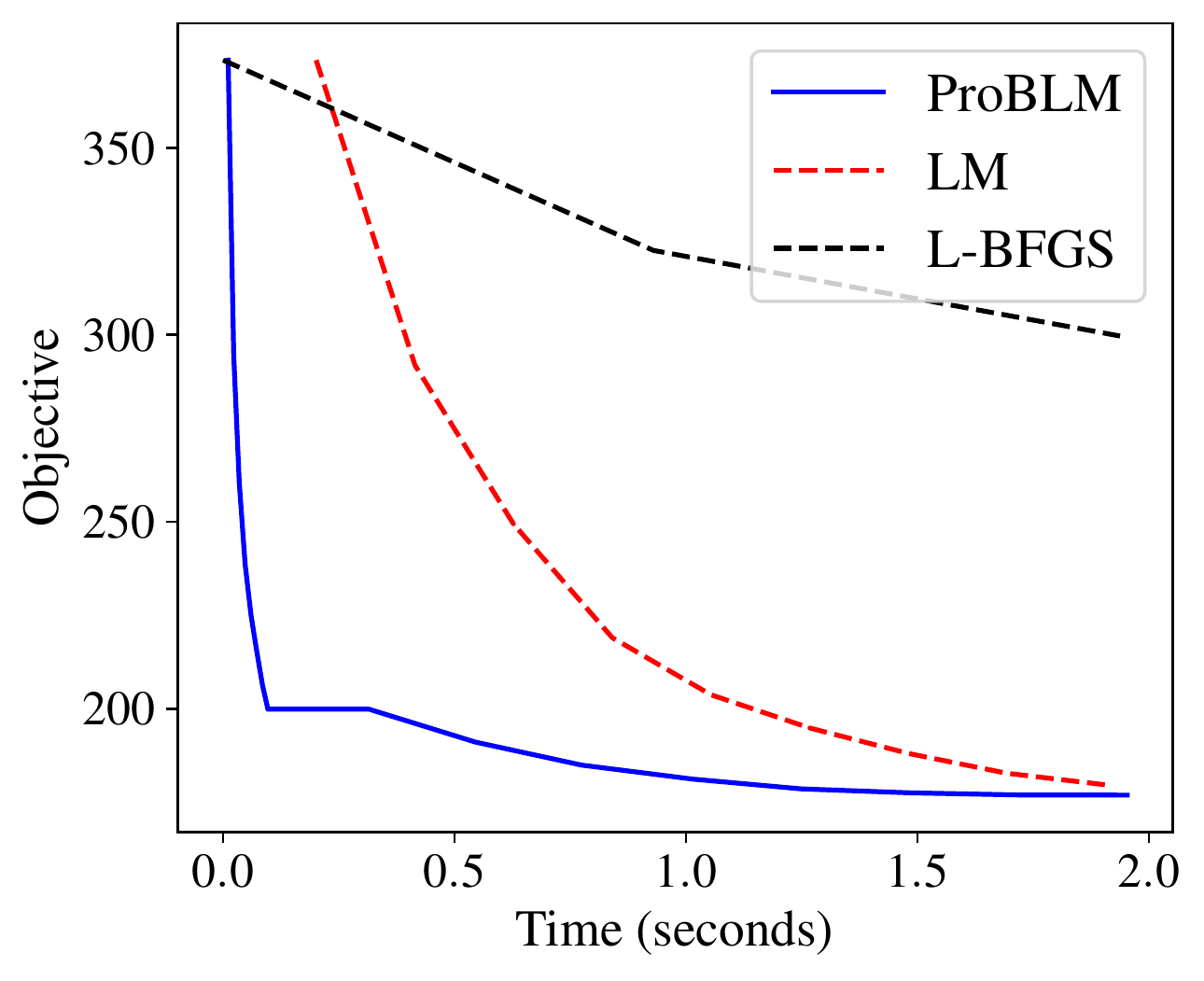}
    \includegraphics[width = 0.32\textwidth]{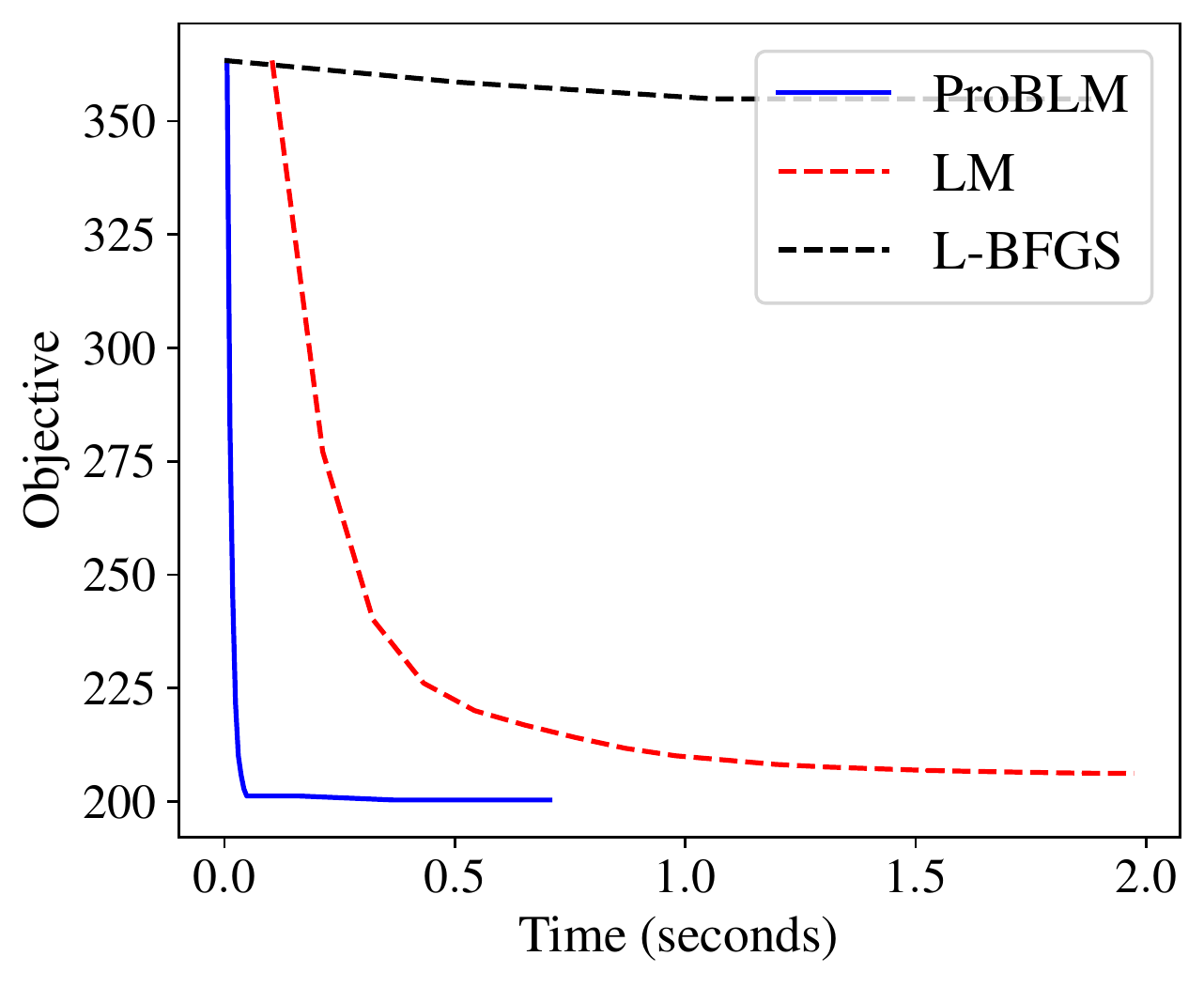}
    \includegraphics[width = 0.32\textwidth]{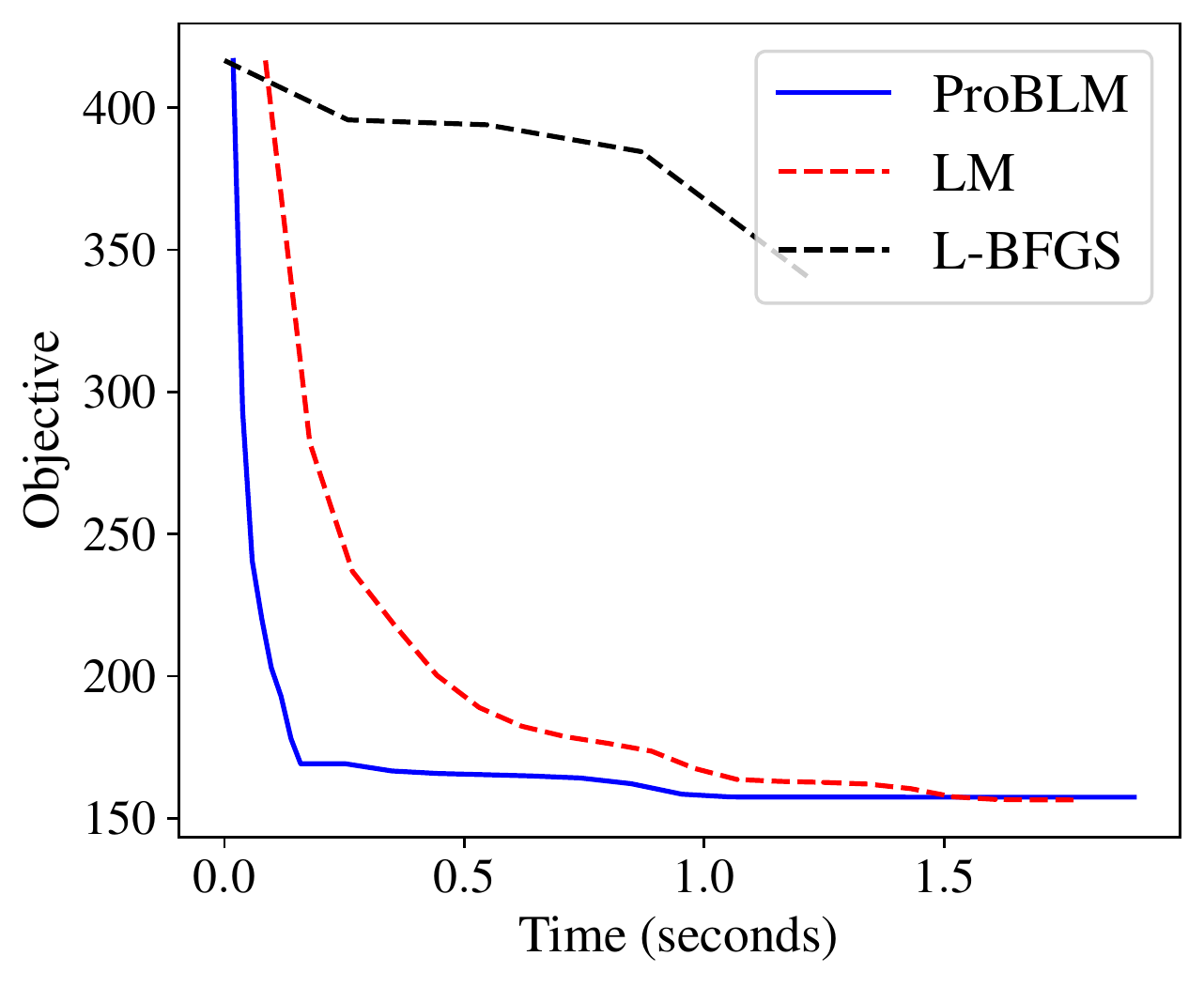}

    \includegraphics[width = 0.32\textwidth]{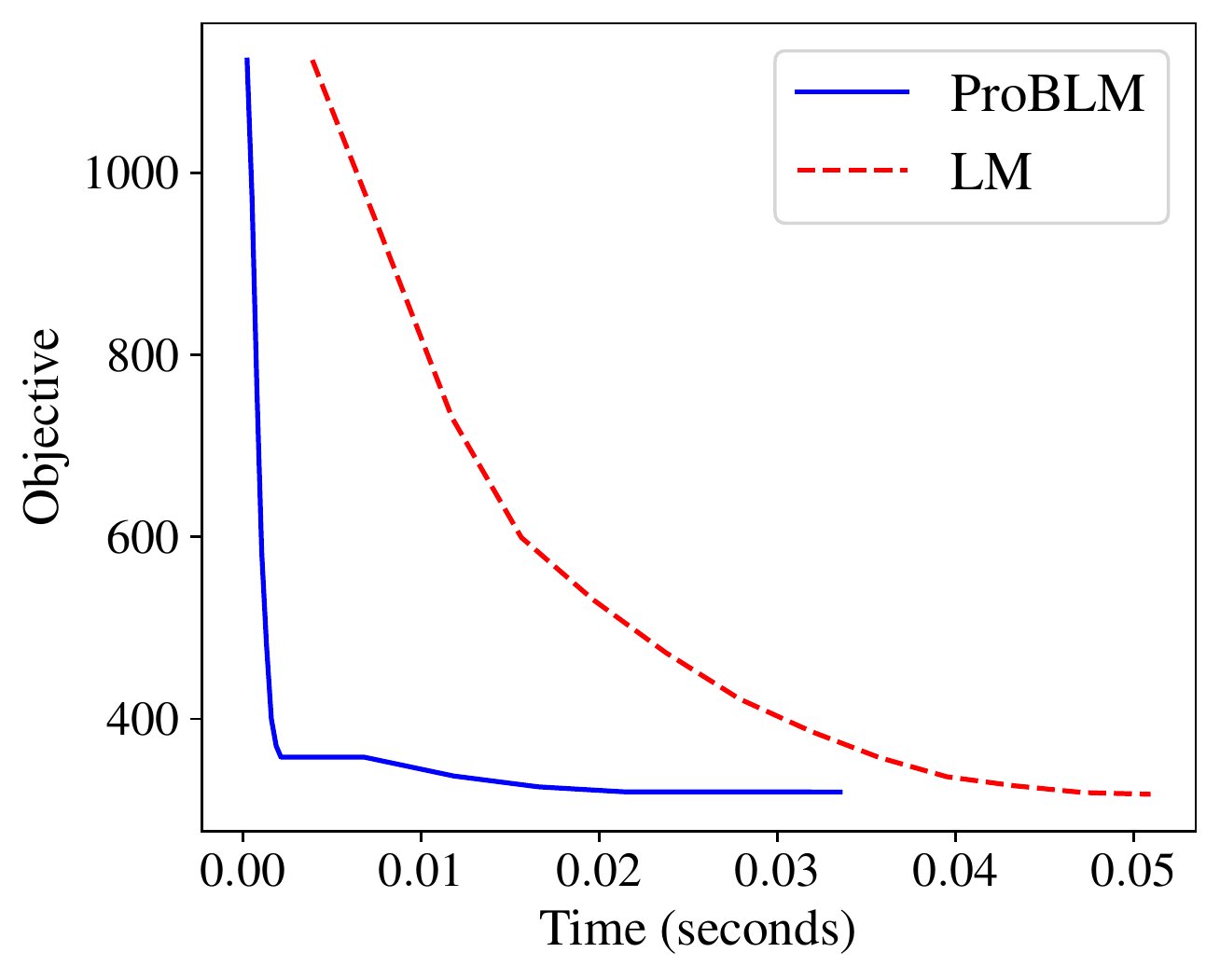}
    \includegraphics[width = 0.32\textwidth]{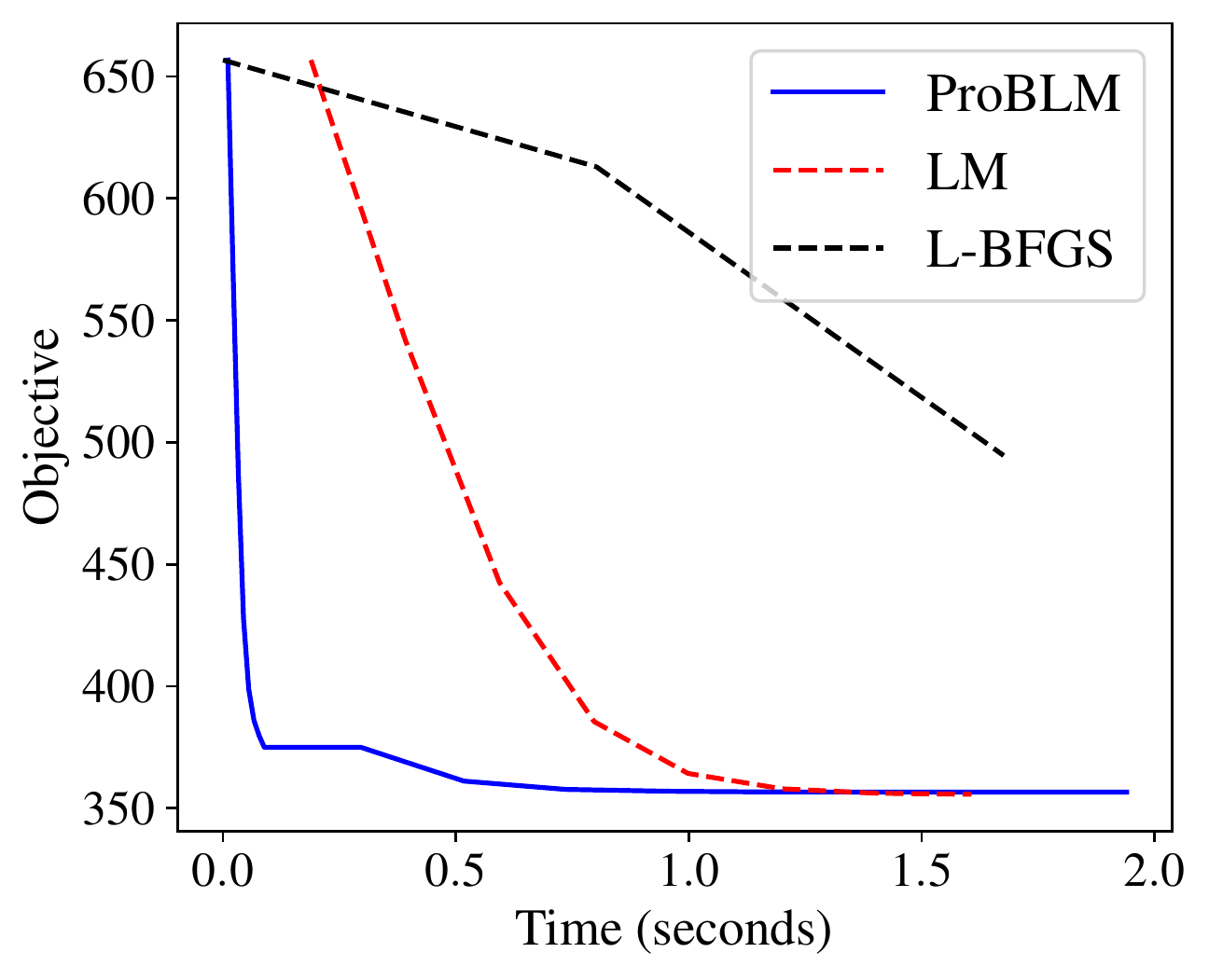}
    \includegraphics[width = 0.32\textwidth]{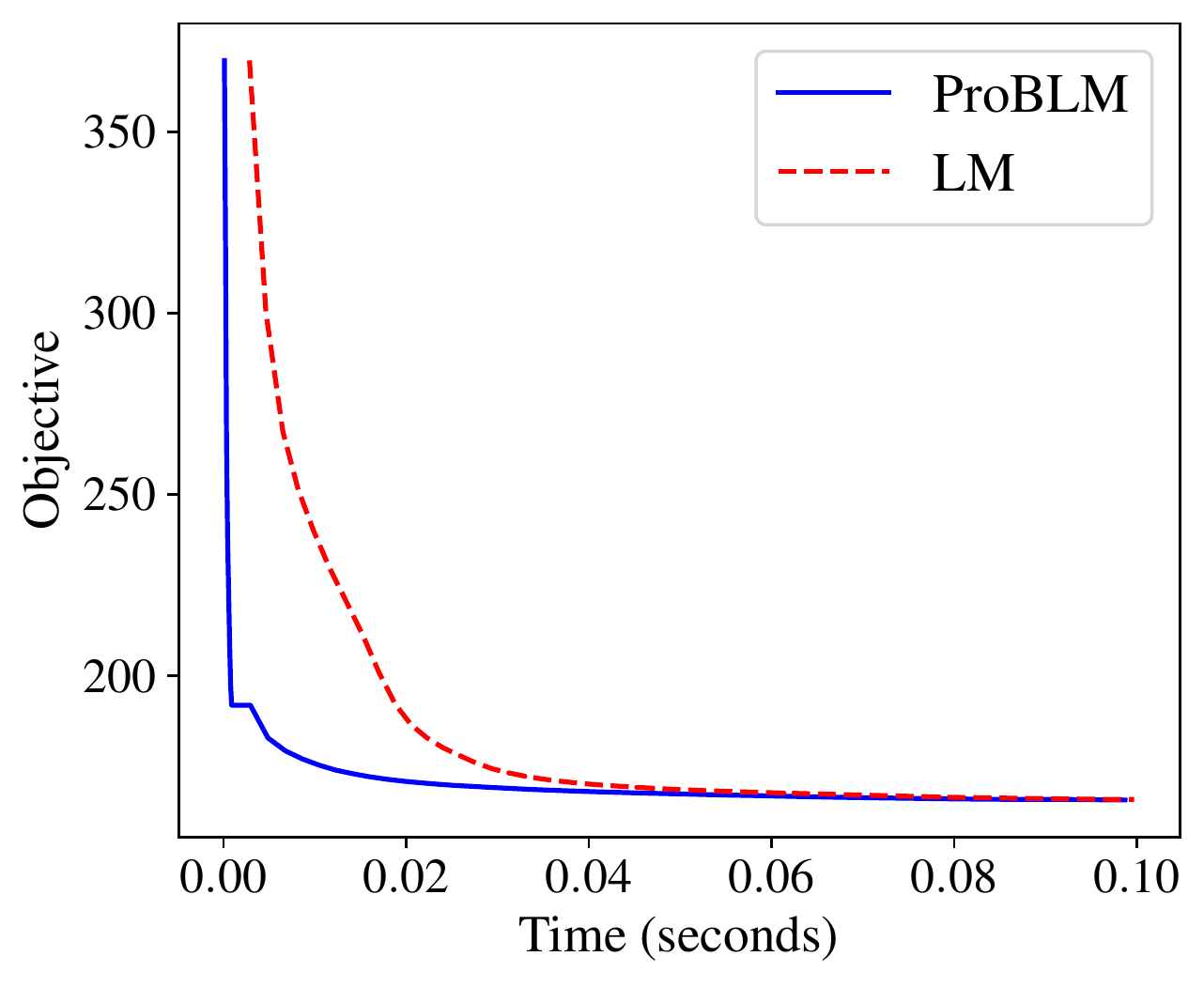}

    \caption{Plots of objective vs. time for our method in comparison with LM and L-BFGS on dense image alignment. The image pairs used are (from left to right, top to bottom): Head, Hartley Building, Union House, Old Classics Wing, Johnson, and Napier Building.}
    \label{fig:dense_homography_results}
\end{figure}

\subsection{Essential Matrix Estimation with Squared Residuals}
High-resolution image pairs from the \emph{``facade"} sequence extracted from the ETH3D dataset are used in this experiment. For each image, we extract SIFT key points, and use nearest neighbor search to get approximately $5000$ putative correspondences per image pair. The key points are normalized using the corresponding intrinsic matrices provided with the dataset. To obtain an outlier-free correspondence set, we run 100 RANSAC iterations on the putative matches to obtain around $2000$ inliers per image pair, which are then fed into the non-linear least squares solvers for refinement. The objective of interest is the total Sampson error induced by all residuals, and we use the parameterisation of~\cite{rosten_2010_improved}.

\begin{figure}
    \centering
    \begin{subfigure}{0.45\textwidth}
        \centering
        \includegraphics[width = 0.99\textwidth]{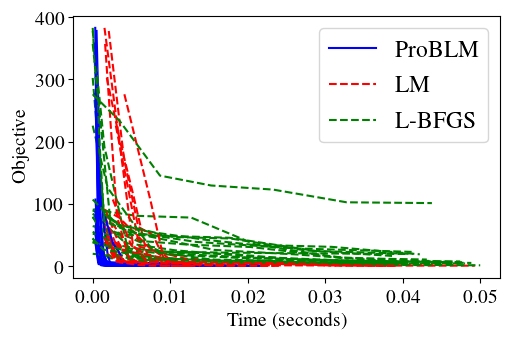}
        \caption{Essential Matrix Estimation}
        \label{subfig:essential_clean}
    \end{subfigure}
    \begin{subfigure}{0.45\textwidth}
        \centering
        \includegraphics[width = 0.99\textwidth]{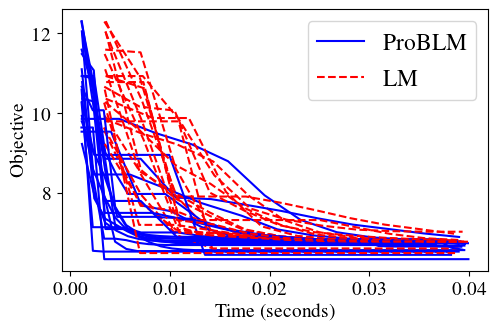}
        \caption{Robust Essential Matrix}
        \label{subfig:essential_robust}
    \end{subfigure}
    \caption{Objectives vs run-time for 20 runs with random initializations for non-robust (left) and robust essential matrix essential matrix estimation (right).}
    \label{fig:essential_results}
\end{figure}

We first evaluate the performance of the 
algorithms on a single pair of images with different random starting points. Fig.~\ref{subfig:essential_clean} shows the objectives versus run time for a single pair of image on $20$ runs, where at the beginning of each run, a random essential matrix is generated and used as starting solution for all methods. Similar to the case of dense image alignment shown in Fig.~\ref{fig:dense_homography_results}, ProBLM demonstrates superior performance throughout all the runs. 

\begin{figure}
    \centering
     \begin{subfigure}{0.45\textwidth}
        \centering
        \includegraphics[width = 0.99\textwidth]{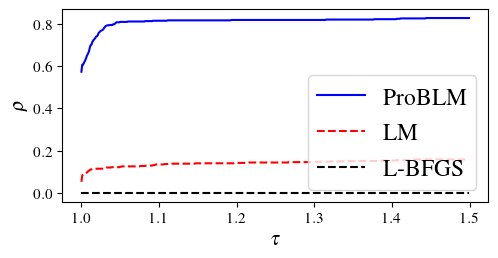}
        \caption{Essential Matrix Estimation}
        \label{subfig:profile_clean}
    \end{subfigure}
    \begin{subfigure}{0.45\textwidth}
        \centering
        \includegraphics[width = 0.99\textwidth]{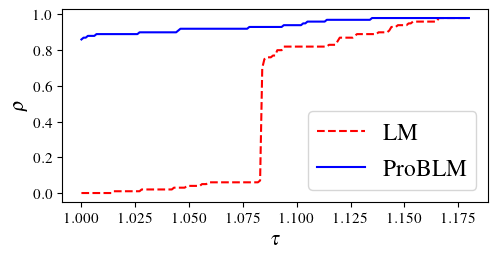}
        \caption{Robust Essential Matrix}
        \label{subfig:profile_robust}
    \end{subfigure}
    
    \caption{Performance profiles for (left): essential matrix estimation with run time budget set to $10$ms, and (right): robust essential matrix estimation with run time budget set to $200$ ms. }    \label{fig:essential_profiles}
\end{figure}

The experiment is repeated for 50 different pairs of images. For each pair, we execute 100 different runs and record their progresses within a run time budget of $10ms$. The results are summarized in Fig.~\ref{subfig:profile_clean}, where we use performance profiles~\cite{dolan2002benchmarking} to visualize the overall performance.  For each image pair, we record the minimum objective $f^*$ obtained across $100$ runs, then measure the percentage of runs (denoted by $\rho$) that achieves the cost of  $\le \tau f^*$ ($\tau \ge 1$) at termination. Fig.~\ref{subfig:profile_clean} shows the results. Observe that within a time budget of $10ms$, a large fraction of ProBLM runs achieve the best solutions, while most LM runs take much longer time to converge. This shows that our method is of great interest for real-time applications.

\subsection{Robust Essential Matrix Fitting}
As introduced earlier, our method is directly applicable to model fitting problems with robust kernels. To demonstrate this, we repeat the essential matrix fitting problem as discussed in the previous section, but we use the set of $5000$ putative correspondences as input. To enforce robustness, we apply the smooth truncated least squares kernel shown in Eq.~\eqref{eq:truncated_lsq}. Graduated Non-convexity~\cite{zach2018descending} with $5$ graduated levels is employed as the optimization framework. At each level (outer loop), our method is used to replace LM and the problem is optimized until convergence before switching to the next level. We compare this traditional approach where LM is used in the nested loop. 
Fig.~\ref{subfig:essential_robust} and~\ref{subfig:profile_robust} show the evolution and performance profile (with the time budget of $200ms$) for this experiment. Similar to the case of clean data, our proposed method outperforms traditional LM by a large margin. A comparison with RANSAC can be found in the supplementary material, where we demonstrate that by applying ProBLM, one achieves comparable solutions to RANSAC within the same amount of run time, which further strengthens the applicability of our method for a wide range of vision problems.

\section{Conclusion}

We propose to accelerate the Levenberg-Marquardt method by utilizing subsampled estimates for the gradient and approximate Hessian information, and by dynamically adjusting the sample size depending on the current progress. Our proposed method has a straightforward convergence guarantee, and we demonstrate superior performance in model fitting tasks relevant in computer vision.

One topic for future research is to investigate in advanced algorithms addressing large-scale and robustified non-linear least-squares problems in order to improve the run-time performance and the quality of the returned solution.

\bibliographystyle{splncs}
\bibliography{main}

\appendix
\noindent
{\Large \textbf{Supplementary Material}}

\section{More Results on Dense Image Alignment}
\begin{figure}[ht]
    \centering
    \includegraphics[width = 0.45\textwidth]{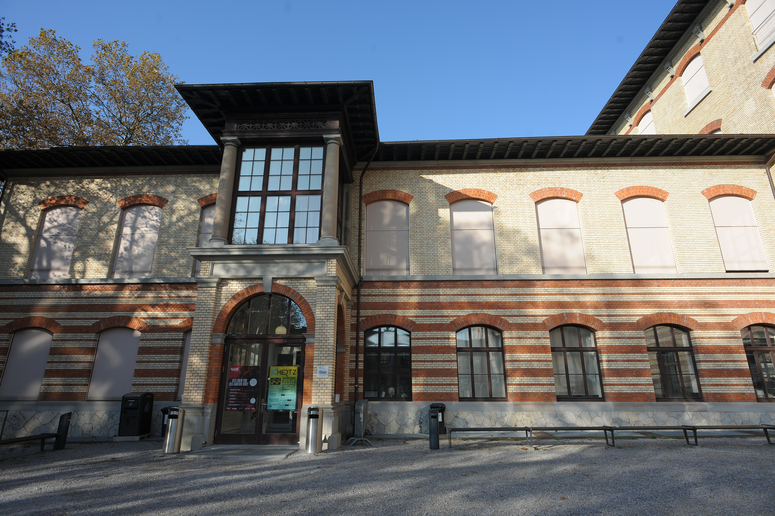}
    \includegraphics[width = 0.45\textwidth]{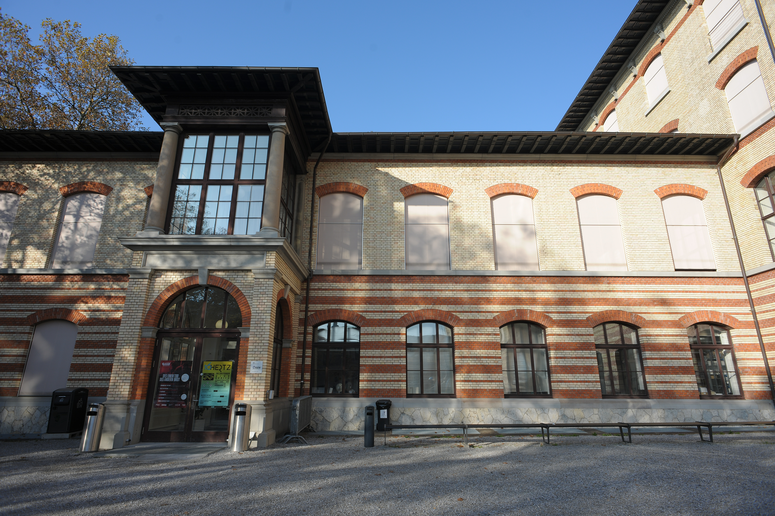}
    \includegraphics[width = 0.8\textwidth]{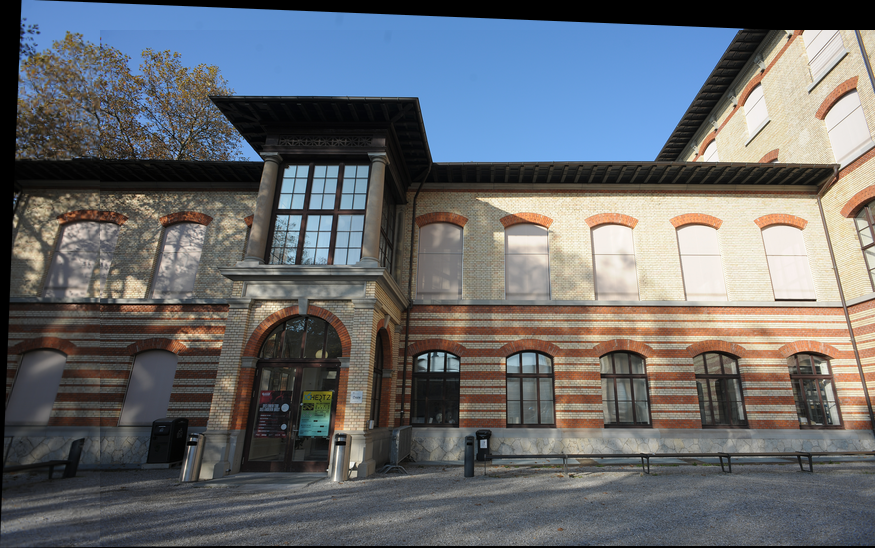}

    \caption{Qualitative results for dense image alignment experiment. Top: Source and target images. Bottom: Alignment result.}
    \label{fig:qualitative_results}
\end{figure}

In this section, we provide additional results on the dense image alignment experiment. Fig.~\ref{fig:dense_homography_results} plots the evolution of 4 more image pairs in the ETH-3D dataset. Note that L-BFGS performs poorly, hence we omit their results. We also show an example of qualitative result in Fig.~\ref{fig:qualitative_results}.
\begin{figure}[ht]
    \centering
    \includegraphics[width = 0.45\textwidth]{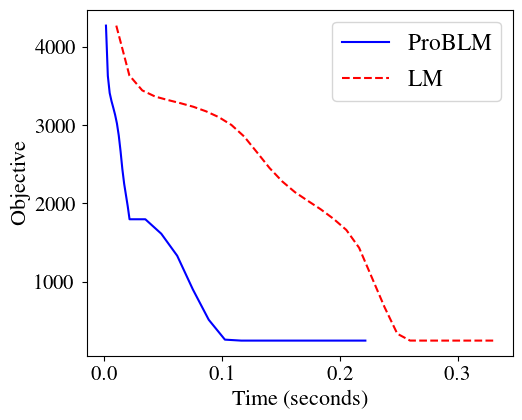}
    \includegraphics[width = 0.45\textwidth]{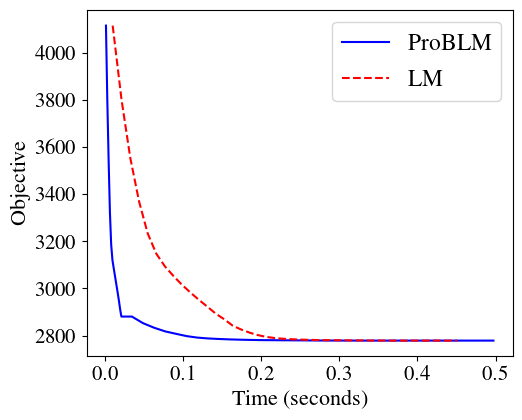}
    \includegraphics[width = 0.45\textwidth]{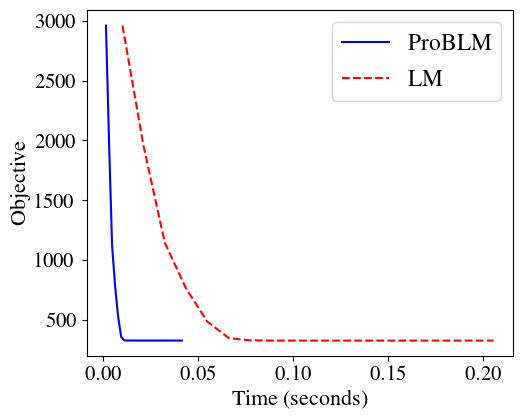}
    \includegraphics[width = 0.45\textwidth]{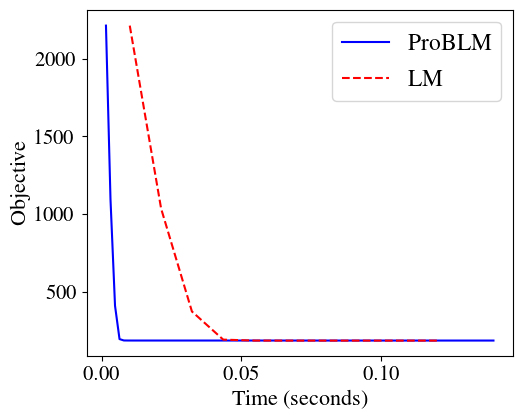}
    
    \caption{Plots of objective vs. time for our method in comparison with LM on dense image alignment.}
    \label{fig:dense_homography_results}
\end{figure}

\section{Weak-Perspective Affine Bundle Adjustment}
\begin{figure}[ht]
    \centering
    \includegraphics[width = 0.45\textwidth]{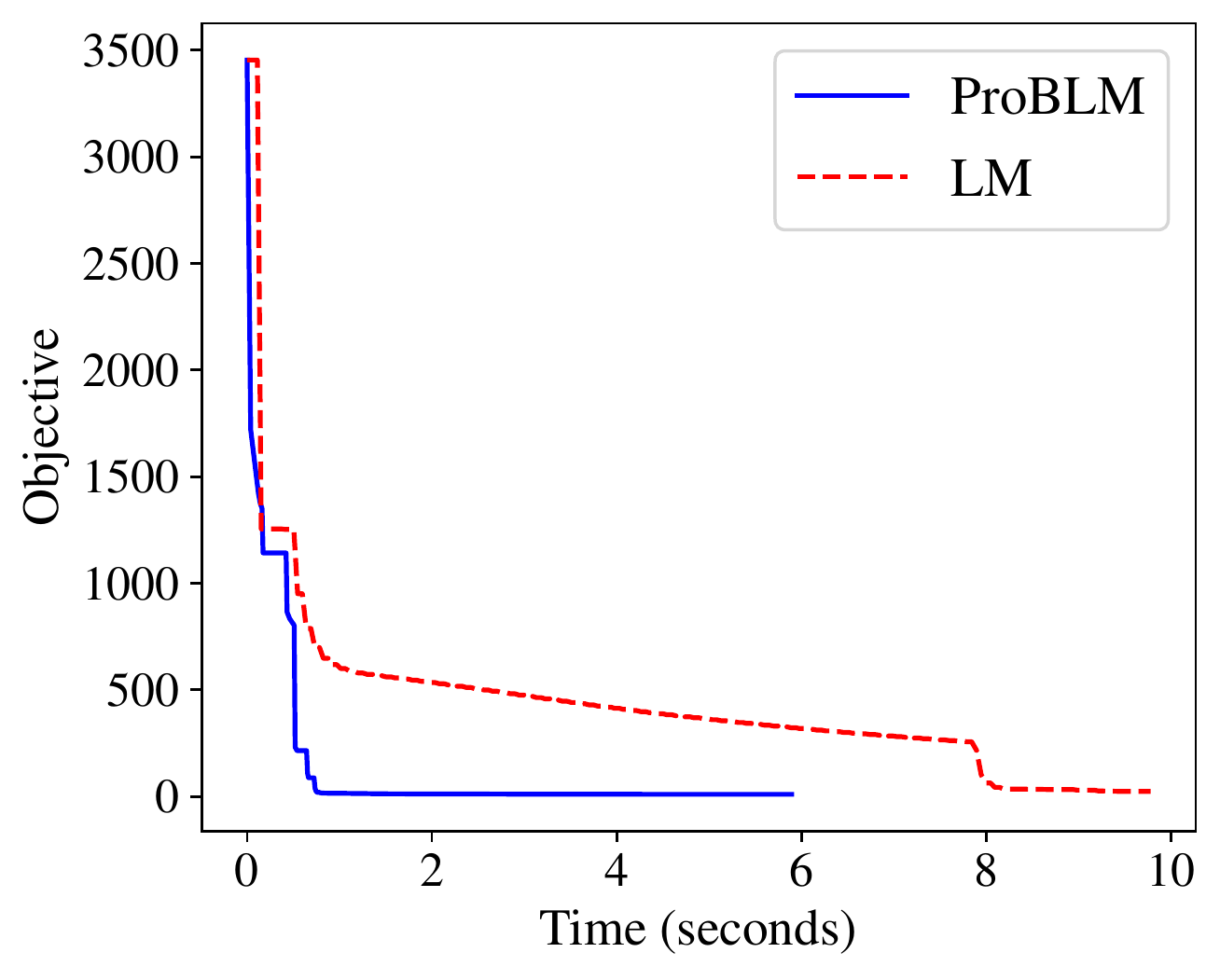}
    \includegraphics[width = 0.45\textwidth]{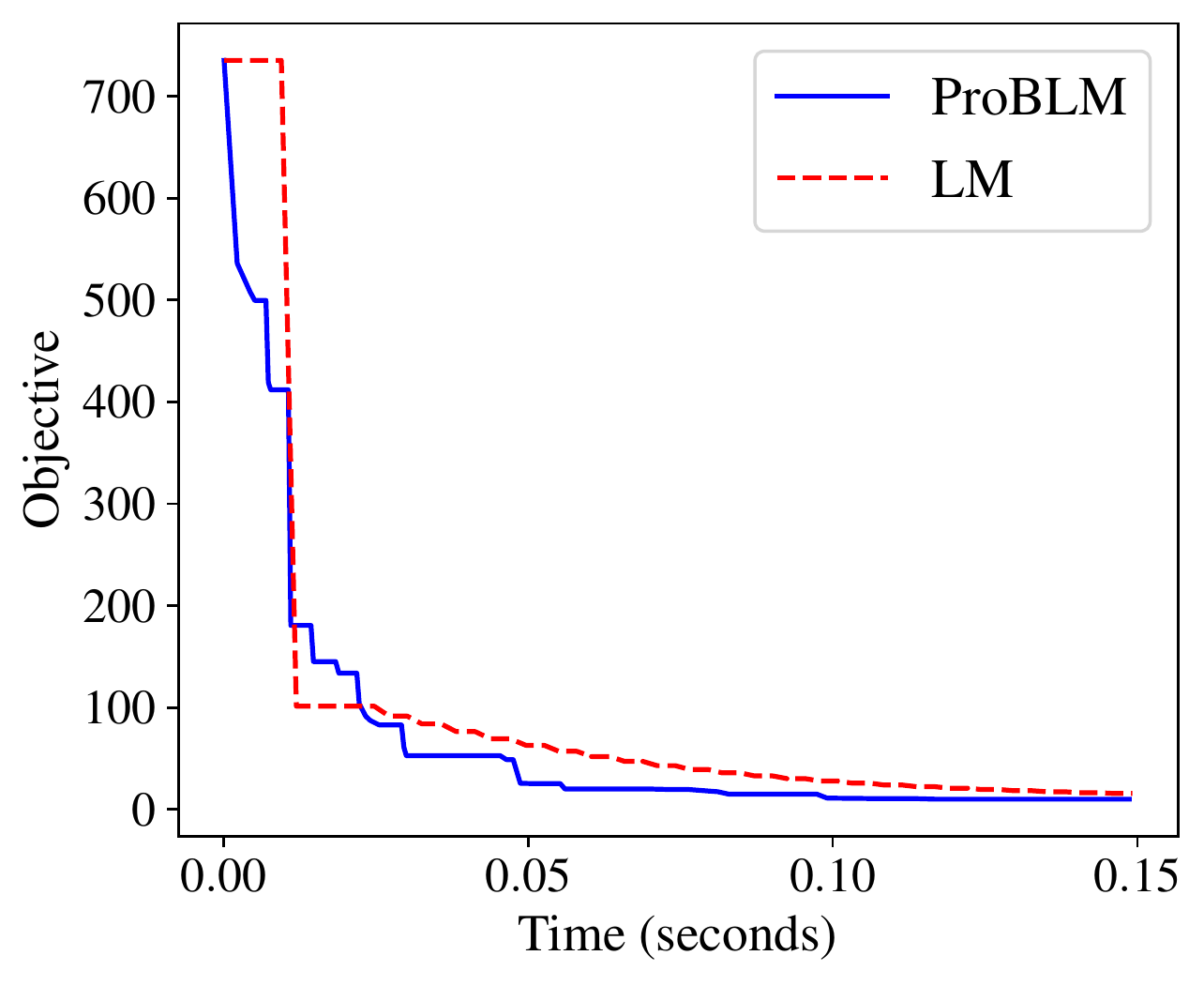}
    \caption{Plots of objective of LM and ProBLM for bundle adjustment. Left: A problem instance with with 3 views. Right: A problem instance with 5 views. }
    \label{fig:bundle}
\end{figure}

We also test the performance of our algorithm on small-scale affine bundle adjustment problems. The objective function can be written as 
\begin{align}
    \min_{\{\bR_i, \bt_i\}, \{\bX_j\}}\sum_{i=1}^{C} \sum_{j=1}^{P} \| \mathbf{\pi}(\bR_{i}\bX_j + \bt_i) - \bm_{ij}\|^2,
\end{align}
where $C$ is the number of cameras and $P$ is the number of 3D points. The parameters $(\bR_i, \bt_i)$ are respectively the rotation and translation that map a 3D point $\bX \in \bbR^3$ from the world coordinate to the coordinate of $i$-th camera, and $\bm_{ij}$ is the 2D projection of the point $\bX_j$ onto the image of camera $i$. Here we use the weak-perspective affine model, i.e.,
\begin{align}
    \mathbf{\pi}([x_1, x_2, x_3]^T) = \left[\frac{x_1}{\tilde{x}_3}, \frac{x_2}{\tilde{x}_3}\right]^T,
\end{align}
where $\tilde{x}_3$ is a fixed average depth. In our experiments, instead of using a single average depth for all 3D points, we associate each point $\bX_j$ with an average depth $\tilde{x}_j$, where $\tilde{x}_j$ is assigned with the initial depth of the point $\bX_j$ and fixed throughout the optimization process. Observe that, with this affine model, when $\{\bR_i\}$ and $\{\bt_i\}$ are fixed, the points can be solved in closed form. Therefore, we employ variable projection~\cite{hyeong2017revisiting} to first solve for the camera parameters, then update the points using standard linear least squares. 
In this experiment, we focus on settings where every 3D point is visible in all cameras, which is a common setting in e.g., several monocular SLAM applications. The ``South Building" dataset from the COLMAP package\footnote{\url{https://demuc.de/colmap/datasets/index.html}} is used, where we extract $3$ adjacent frames and all 3D points that are visible in all extracted frames, resulting in a problem instance containing $802$ points in 3D and $2406$ measurements. 
Fig.~\ref{fig:bundle} (left) plots the objective for conventional LM and our method. Observe that ProBLM also offers favorable results. The same experiment is repeated for $5$ views with $212$ 3D points, and the results is plotted in Fig.~\ref{fig:bundle} (right). It can also be seen that ProBLM converges faster (although LM has comparable performance in this case because the number of 3D points is smaller than the case of three views shown in the left figure).

\section{Algorithm for Relaxed Condition}
\begin{figure}[ht]
    \centering
    \includegraphics[width = 0.45\textwidth]{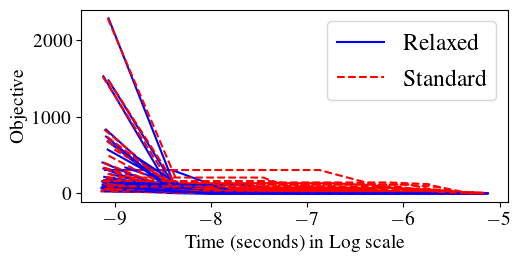}
    \includegraphics[width = 0.45\textwidth]{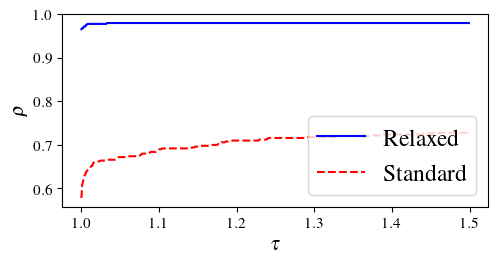}
    \caption{Comparison between the two conditions. Left: Plots of objective vs. run time for 20 runs. Right: Performance profile with a 10ms time budget, summarized over 500 runs. }
    \label{fig:compare_conditions}
\end{figure}

Algorithm~\ref{alg:RePROBLM} describes the relaxed version of our method (using Eq.\ (18)). In Fig.~\ref{fig:compare_conditions}, we compare the performance of the standard condition (Eq.\ (13)) and relaxed condition (Eq.\ (18)). As can be seen, the relaxed condition generally offers better performance.
\begin{algorithm}[t]
  \centering
  \caption{Relaxed ProBLM}
\label{alg:RePROBLM}
\begin{algorithmic}[1]                   
  \REQUIRE Initial solution $\btheta^{(0)}$, initial batch size $K_0$, maximum iterations \texttt{max\_iter}
  \REQUIRE Confidence level $\delta \in (0,1)$, margin parameter $\alpha \in [0,1)$, $\eta$
  \STATE Randomly reshuffle the residuals $\{f_i\}$
  \STATE Initialization: $t \gets 0$, $K \gets K_0$, $t_0 \gets 0$.
  \WHILE{$t<$ \texttt{max\_iter} and a convergence criterion is not met}
  \STATE $\cS^{(t)} \gets \{1, \dotsc, K\}$
  \STATE Compute $\bg_{\cS^{(t)}}$ and $\bH_{\cS^{(t)}}$
  \begin{align}
    {\bg}_{\cS^{(t)}} \coloneqq \sum\nolimits_{i \in \cS^{(t)}} \bJ^{(t)}_i \br_i^{(t)} & &  {\bH}_{\cS^{(t)}} \coloneqq \sum\nolimits_{i \in \cS^{(t)}} (\bJ^{(t)}_i)^T \bJ^{(t)}_i.
  \end{align}
  and solve
  \begin{align}
    \Delta {\btheta}^{(t)} \gets \left( {\bH}_{\cS^{(t)}} + \lambda \bI \right)^{-1}{\bg}_{\cS^{(t)}} & & \btheta^{(t+1)} \gets \btheta^{(t)} + \Delta {\btheta}^{(t)}
  \end{align}
  \IF{$f_i(\btheta^{(t+1)}) - f_i(\btheta^{(t)} \ge 0$}
  \STATE $\btheta^{(t+1)} \gets \btheta^{(t)}$, and $\lambda \gets 10\, \lambda$ \hfill $\vartriangleright$ Failure step
  \ELSE
  \STATE Determine current lower and upper bounds $a$ and $b$, and set
  \begin{align}
      U_K^{(t_0, t+1)}\gets \sum\nolimits_{i\in \cS^{(t)}} \max\left\{ a, \left( f_i(\btheta^{(t+1)}) - f_i(\btheta^{(t_0)}) \right) \right\}.
  \end{align}
  \STATE $p \gets $ Random number between $0$ and $1$.
    \IF{ {$U_K$ satisfies Eq.~(18)} \OR {$p \le \eta$ } } 
    \STATE $\lambda \gets \lambda/10$ \hfill $\vartriangleright$ Success step
    \ELSE
      \STATE $\btheta^{(t+1)} \gets \btheta^{(t)}$ and increase $K$ using Eq.~(16). (with $S_K$ replaced by $U_K$)
      \STATE $t_0 \gets (t+1)$
    \ENDIF
  \ENDIF
  \STATE $t \gets t+1$
  \ENDWHILE
  \RETURN $\btheta^{(t)}$
\end{algorithmic}
\end{algorithm}

\section {RANSAC vs. Robustified ProBLM}
\begin{figure}[ht]
    \centering
    \includegraphics[width = 0.80\textwidth]{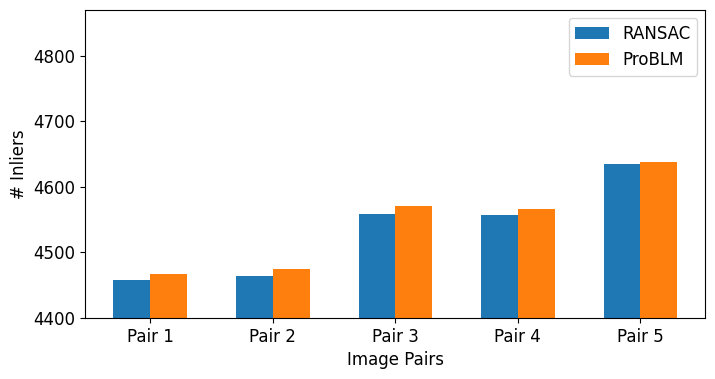}
    
    \caption{Number of inliers obtained after a fixed time budget of RANSAC and robustified ProBLM.}
    \label{fig:inls}
\end{figure}

In Fig.~\ref{fig:inls}, we plot the number of inliers obtained using RANSAC and robustified ProBLM of 5 randomly chosen image pairs from the ETH3D dataset. For every image pair, each method is run 10 times with a fixed time budget (ProBLM is initialized with random initializations), and the reported results are averaged over 10 runs. As can be observed from Fig.~\ref{fig:inls}, within the same run-time budget, ProBLM achieves competitive number of inliers without the need of RANSAC. This suggests that our method has the potential of directly fitting the model without requiring RANSAC as a pre-processing step, which is highly relevant for many real-time applications.

\end{document}